\documentclass{article}
\usepackage{spconf,amsmath,graphicx,hyperref}
\usepackage{booktabs,makecell,pifont,graphicx,xcolor,adjustbox}
\newcommand{\ok}{\textcolor{green!40!black}{\ding{51}}}  
\newcommand{\bad}{\textcolor{red!50!black}{\ding{55}}}  
\newcommand{\Model}[1]{\textsf{#1}}

\title{Beyond Pixels: A Vector-to-Graph Framework for Reliable Schematic Auditing}
\name{
  Chengwei Ma$^{1}$,
  Zhen Tian$^{1}$,
  Zhou Zhou$^{1}$,
  Zhixian Xu$^{2}$,
  Xiaowei Zhu$^{2}$,
  Xia Hua$^{3}$,
  Si Shi$^{1}$,
  F. Richard Yu$^{4}$
\thanks{Corresponding author: Zhen Tian (tianzhen@gml.ac.cn)}
}

\address{
  $^{1}$ Guangdong Laboratory of Artificial Intelligence and Digital Economy (SZ), Shenzhen, China \\
  $^{2}$ Guangdong Power Grid Co., Ltd., Yangjiang Power Supply Bureau, Yangjiang, China \\
  $^{3}$ Shanghai University, Shanghai, China \\
  $^{4}$ Carleton University, Ottawa, Canada
}

\begin{document}
%\ninept
%
\maketitle
\begin{abstract}
Multimodal Large Language Models (MLLMs) have shown remarkable progress in visual understanding, yet they suffer from a critical limitation: structural blindness. Even state-of-the-art models fail to capture topology and symbolic logic in engineering schematics, as their pixel-driven paradigm discards the explicit vector-defined relations needed for reasoning. To overcome this, we propose a Vector-to-Graph (V2G) pipeline that converts CAD diagrams into property graphs where nodes represent components and edges encode connectivity, making structural dependencies explicit and machine-auditable. On a diagnostic benchmark of electrical compliance checks, V2G yields large accuracy gains across all error categories, while leading MLLMs remain near chance level. These results highlight the systemic inadequacy of pixel-based methods and demonstrate that structure-aware representations provide a reliable path toward practical deployment of multimodal AI in engineering domains. To facilitate further research, we release our benchmark and implementation at https://github.com/gm-embodied/V2G-Audit.
\end{abstract}


\section{Introduction}
\label{sec:intro}

Multimodal large language models (MLLMs)  have achieved strong progress in perception, yet they suffer from \emph{structural blindness}: difficulty with spatial relations, counting, and connectivity. Recent studies show that MLLMs often exploit priors rather than genuinely parsing diagram structure~\cite{hou2024vision}, and perform unreliably on spatial reasoning~\cite{wang2024picture}. Such limitations are critical in schematic auditing, where correctness depends on topological and symbolic constraints rather than visual appearance.

\begin{figure}[!t]
\centering
\includegraphics[width=0.4\textwidth]{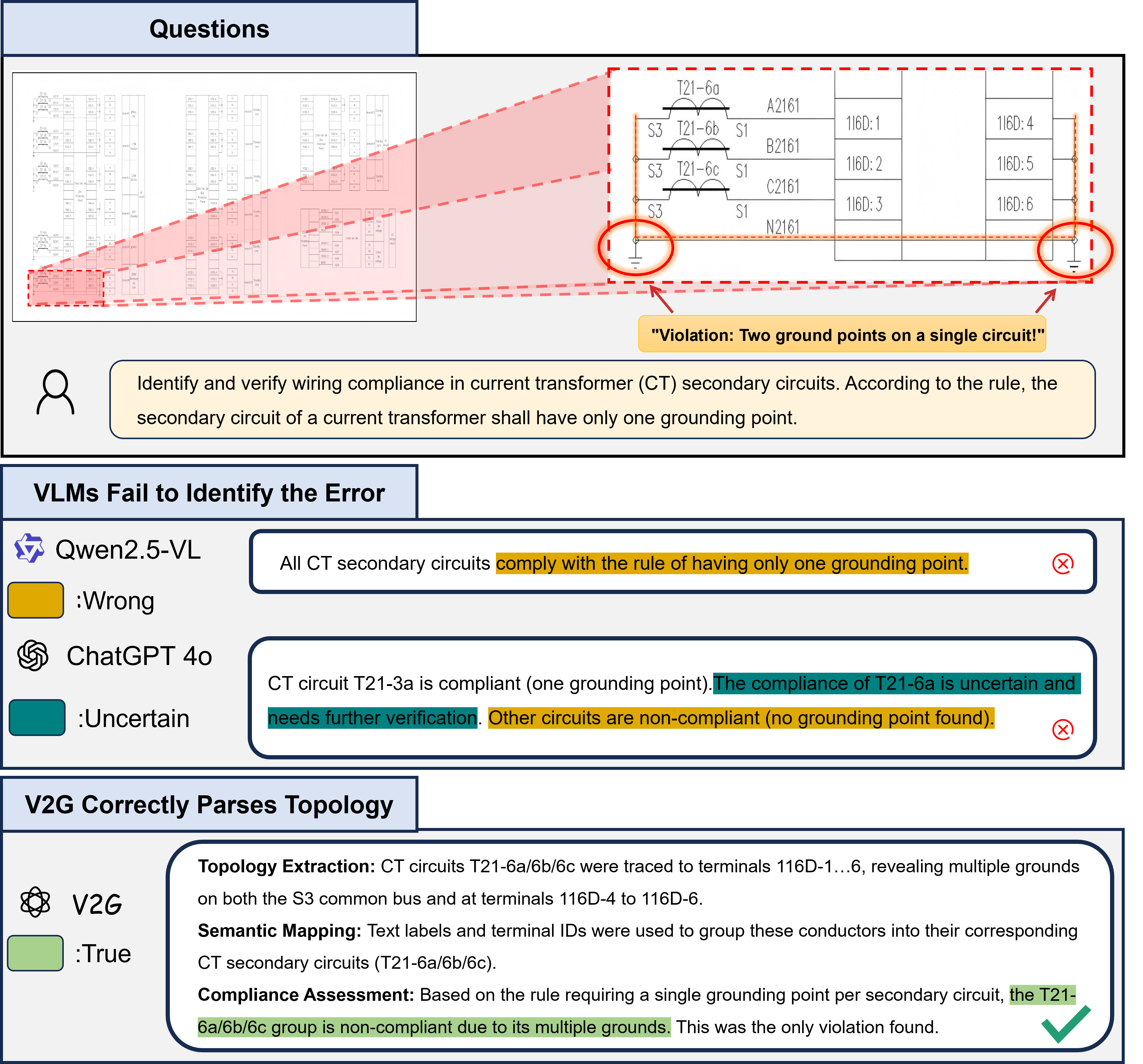} 
\caption{State-of-the-art MLLM fails to ensure single-point CT grounding, exemplifying \emph{structural blindness}.}
\label{fig:teaser}
\end{figure}

High-resolution pipelines—tiling, native-resolution encoders, or salient-token selection—improve pixel acuity but remain confined to the \emph{pixel domain}. Even advanced systems that restore global coherence, such as HiRes-LLaVA~\cite{huang2024hiresllava} or 4K scaling~\cite{shi2025scaling}, still operate on rasterized inputs. By contrast, object-centric approaches (e.g., OCRA~\cite{webb2023objectcentric} and \cite{locatello2020object}) demonstrate the benefits of decomposing scenes into entities and relations. For Computer-Aided Design (CAD) diagrams, vector grounding is essential; VDLM~\cite{wang2024VDLM} shows that converting vector graphics into symbolic descriptions enables precise structural reasoning.

Graph Signal Processing (GSP) offers principled operators for structure-aware verification on irregular domains~\cite{shuman2013signal,ortega2018graph}, with recent engineering applications confirming its effectiveness~\cite{liu2024gift}. Diagnostic datasets such as CLEVR~\cite{johnson2017clevr} and CIRCUIT~\cite{skelic2024circuit} further expose persistent failures of current models on compositional and circuit reasoning.

\begin{figure*}[htb]
\centering
\includegraphics[width=0.85\linewidth]{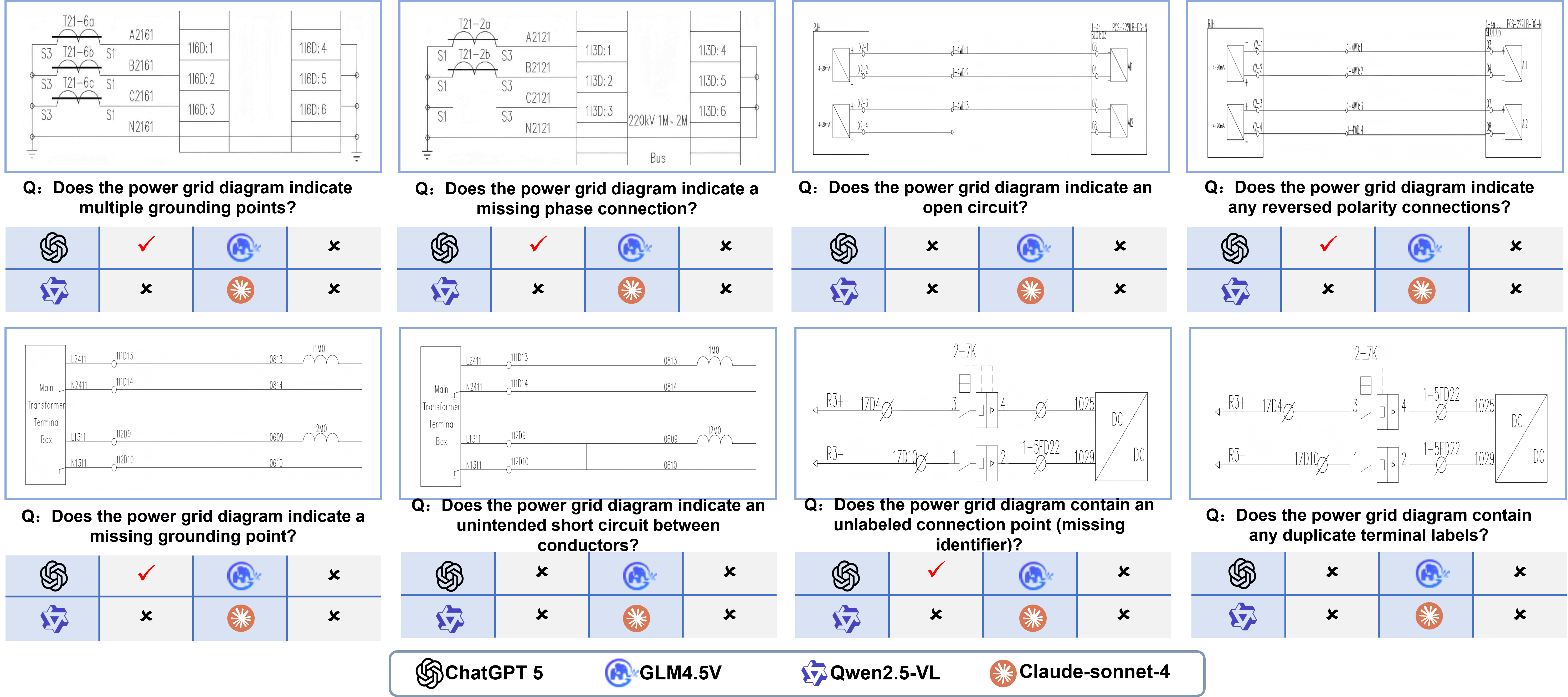} 
\caption{Representative compliance tasks used to expose structural blindness. MLLMs fail consistently across grounding, wiring, and labeling checks, revealing inability to reason over schematic topology.
}
\label{fig:failure_case}
\end{figure*}

To address these gaps, we propose a \textbf{Vector-to-Graph (V2G)} framework for schematic auditing. Instead of raster inputs, CAD schematics are parsed into property graphs where nodes denote components, edges encode connectivity, and attributes capture polarity, grounding, and terminal IDs. Compliance rules are then verified by deterministic GSP operators. This paradigm preserves relational information lost in pixel-based processing, bridges multimodal planning with symbolic verification, and provides a structure-aware alternative that mitigates the structural blindness of modern MLLMs.

\section{Exposing Structural Blindness}
\label{sec:problem}

To demonstrate the limitations of current MLLMs, we design a targeted diagnostic probe that isolates \emph{topological reasoning}, inspired by earlier structured reasoning benchmarks such as CLEVR~\cite{johnson2017clevr}, SpatialEval~\cite{wang2024picture}, and recent diagram evaluations~\cite{hou2024vision,wang2024VDLM}. Our goal is not exhaustive coverage but a clear, domain-specific demonstration of the failure modes that arise when connectivity must be inferred.

\textbf{Case Study: Multi-Point Grounding.}  
We begin with a fundamental compliance rule in electrical engineering: a Current Transformer (CT) secondary must connect to a ground symbol at exactly one point. The visual input is unambiguous and the required logic deterministic. When tested zero-shot, leading high-resolution MLLMs consistently failed this task (Fig.~\ref{fig:teaser}). While they correctly detected components such as ``current transformers'' and ``grounding symbols,'' none could reliably determine that two ground points were connected to the same circuit—often producing confident but incorrect answers, or hedging with multiple ambiguous regions. 

\textbf{Generalization to Other Tasks.}  
We extend the probe to eight representative compliance checks across three categories: grounding (multi-point, missing), wiring (open circuit, polarity reversal, short circuit, missing phase), and labeling (unlabeled or duplicate terminal IDs). Results are consistent: across all cases, every tested model committed errors and none reached stable correctness (Fig.~\ref{fig:failure_case}). These patterns echo findings from CIRCUIT~\cite{skelic2024circuit}, which exposed systematic failures of MLLMs on circuit reasoning.

\textbf{Implications.}  
These failures exemplify \emph{structural blindness}: MLLMs act as ``bag-of-objects'' recognizers, lacking the ability to recover the graph structure implicit in vector-defined schematics. Even with near-perfect pixel-level perception, they fail to enforce basic relational rules. This limitation, also emphasized in studies of diagrammatic and spatial reasoning~\cite{tong2024eyes,kamath2023up}, underscores the need for structure-aware alternatives. In this work we pursue such a path, integrating property-graph representations with graph signal processing~\cite{shuman2013signal,liu2024gift}.

\section{Proposed Framework: Vector-to-Graph with Graph-Level Verification}
\label{sec:method}

\begin{figure}[!t]
    \centering
    \includegraphics[width=0.9\linewidth]{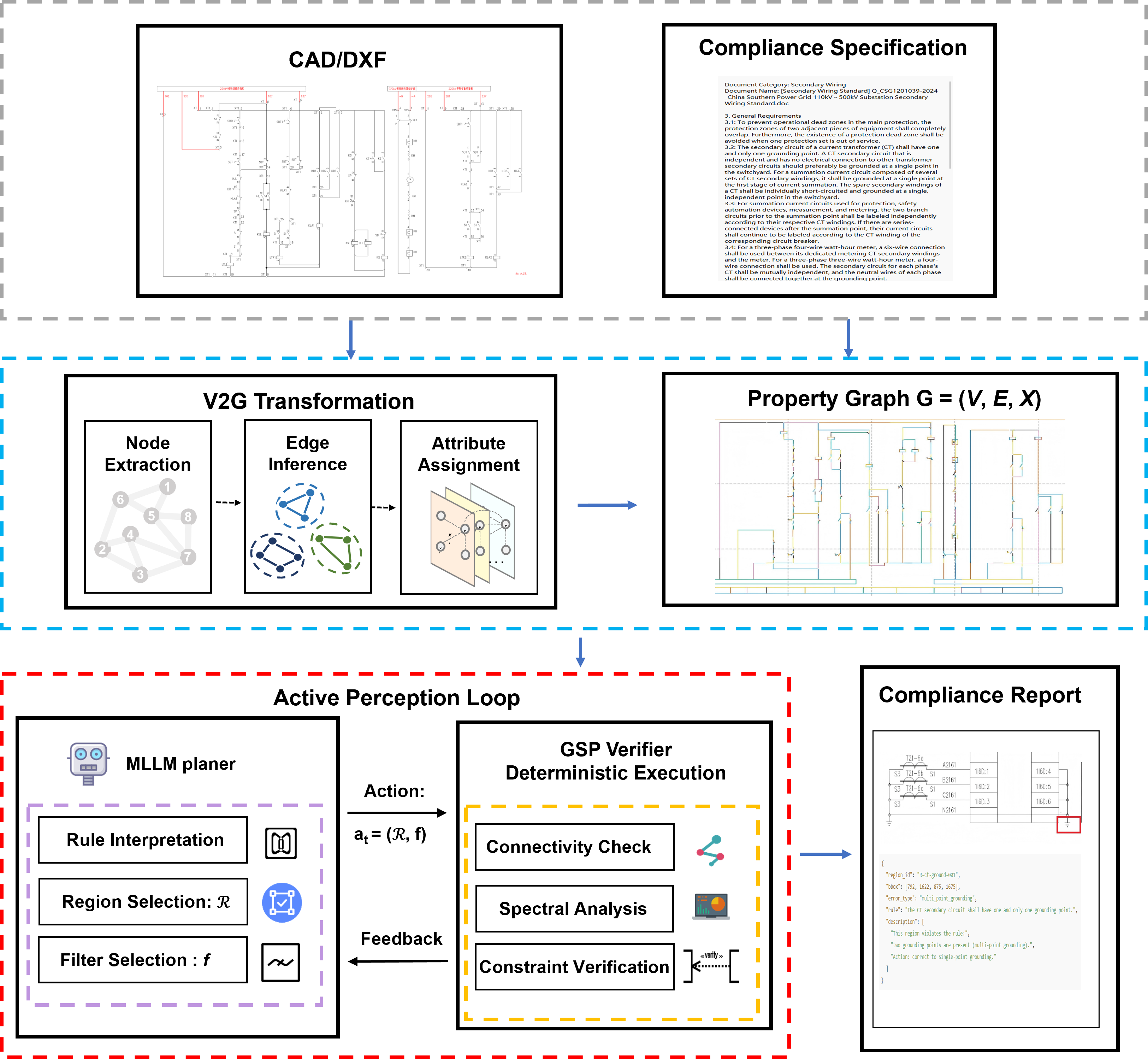} 
    \caption{Proposed V2G-based auditing framework.}
    \label{fig:framework}
\end{figure}

Pixel-based methods suffer from ``Structural Blindness'': they detect symbols but fail to reason about topology~\cite{hou2024vision,wang2024picture}. We address this gap with a Vector-to-Graph (V2G) framework that converts CAD schematics into a property graph and integrates a Large Multimodal Language Model (MLLM) planner with deterministic Graph Signal Processing (GSP) verification (Fig.~\ref{fig:framework}).

% ================= Framework Figure =================
% ====================================================

\textbf{Stage 1: V2G Transformation.}  
We parse CAD schematics with the \texttt{ezdxf} library to obtain low-level primitives $\mathcal{P}=\{p_i\}$ (\texttt{LINE}, \texttt{ARC}, \texttt{TEXT}, \texttt{INSERT}, etc.). Instead of manually defining rules, we employ an LLM-driven pipeline that sequentially extracts nodes, edges, and attributes, analogous to knowledge graph construction. First, component \emph{nodes} are identified by grouping primitives around standardized blocks (\texttt{INSERT}) or recognized patterns (e.g., current transformers, breakers, ground symbols). Second, electrical \emph{edges} are inferred by combining geometric heuristics with LLM interpretation: two entities are connected if their endpoints fall within tolerance $\tau$, unless filtered out as visual crossings. Finally, textual annotations and block attributes are mapped into node \emph{attributes} $X$, including terminal IDs, polarity, or grounding type. The resulting property graph $G=(V,E,X)$ preserves both symbolic semantics and explicit topology~\cite{wang2024VDLM,webb2023objectcentric}.

\textbf{Stage 2: MLLM Planner.}  
Compliance rules $\Phi$ are expressed in natural language. Using structured prompting, the MLLM interprets each rule into a query of the form
\[
a_t = (\mathcal{R}_t, f_t), \quad \mathcal{R}_t \subseteq G, \; f_t \in \mathcal{F},
\]
where $\mathcal{R}_t$ is a relevant subgraph of $G$ and $f_t$ a verifier function from library $\mathcal{F}$. This process consists of three steps. (i) \emph{Rule Interpretation}: the MLLM parses $\phi \in \Phi$ into structured JSON commands. For example, the rule ``Every CT secondary must connect to exactly one ground'' is mapped to  
\begin{verbatim}
{"region":"CT_secondary",
 "function":"check_grounding_uniqueness"}
\end{verbatim}
(ii) \emph{Region Selection}: keywords in the rule (e.g., ``grounding symbol'') are aligned with node types or attributes in $G$, yielding the candidate subgraph $\mathcal{R}_t$. (iii) \emph{Function Selection}: the MLLM retrieves the appropriate verification operator $f_t$ (e.g., connectivity check, attribute consistency, spectral analysis). The structured query is then executed by the deterministic GSP verifier, and feedback is integrated for subsequent iterations.

\textbf{Stage 3: GSP Verifier.}  
For a selected subgraph $\mathcal{R}_t=(V_t,E_t,X_t)$, we define adjacency $A_t$, degree $D_t$, and the Laplacian $L_t=D_t-A_t$. Connectivity is verified using spectral graph tools~\cite{chung1997spectral,vonLuxburg2007tutorial}: the number of connected components equals the multiplicity of the zero eigenvalue, $c=\text{mult}_{\lambda=0}(L_t)$, or equivalently $\mathrm{rank}(L_t)=|V_t|-c$. From a graph signal processing (GSP) perspective, node attributes are treated as signals over $V_t$, and verification functions apply operators induced by $A_t$ or $L_t$~\cite{shuman2013signal,ortega2018graph,sandryhaila2013dsp,liu2024gift}. Grounding uniqueness is tested by
\[
g(\mathcal{R}_t)=\sum_{v\in V_t}\mathbf{1}[\text{type}(v)=\text{Ground}],
\]
where $g=1$ indicates compliance, $g=0$ missing grounding, and $g>1$ multiple groundings. Polarity and phase consistency are enforced through attribute constraints
\[
c_{\text{attr}}(\mathcal{R}_t)=\prod_{(u,v)\in E_t}\mathbf{1}[x_u^{(\text{attr})}=x_v^{(\text{attr})}],
\]
ensuring matched labels across connected components. Wiring loops and abnormal circuits are identified by spectral features of $L_t$, e.g., eigenvalue multiplicity or the cycle number $\beta=|E_t|-|V_t|+c$, and can be further localized via spectral wavelet operators $g(L_t)$~\cite{hammond2011wavelets}. Each verification function returns $o_t\in\{0,1\}$, providing deterministic and auditable compliance outcomes.

\textbf{Stage 4: Compliance Report.}  
Outputs are aggregated as
\[
O=\{(\phi_j,o_j)\}_{j=1}^{|\Phi|},
\]
where each pair indicates whether rule $\phi_j$ is satisfied. Results are exported as structured JSON with violation flags and summarized in natural language for human auditors.

\textbf{Intropy Analysis} Intropy quantifies intelligence as adaptive efficiency, $\mathrm{d}L = \delta S/{R}$, where $\delta S$ is meaningful discrepancy reduction and $R$ denotes internal resistance such as uncertainty or representation mismatch \cite{Yu2026Intropy,Ren2025Intelligence}. Pixel-based schematic understanding exhibits low Intropy due to high structural resistance: topological relations and symbolic constraints remain implicit. In contrast, the proposed Vector-to-Graph framework reduces resistance by explicitly encoding schematic structure as graphs. Graph-based reasoning and GSP verification increase $\delta S$ while lowering $R$,
yielding more efficient, robust, and interpretable schematic auditing.

\section{EXPERIMENTS}
\label{sec:experiments}

\subsection{Experimental Setup}
\label{ssec:setup}
\textbf{Benchmark.}  
The benchmark contains $60$ unique base cases collected from \emph{real engineering schematics} provided by a regional power utility, covering three compliance categories: \emph{connection labeling} (missing/duplicate/misaligned terminal IDs), \emph{grounding} (missing/multi-point/incorrect location), and \emph{wiring} (open circuit, polarity reversal, inter-circuit short, missing phase). Each base case is further augmented with 10--20 rotation, translation, and mild scale/noise variants that preserve electrical semantics, yielding a total of approximately $N{\approx}900$ test instances. To avoid artificially inflating the dataset size, evaluation aggregates results at the \emph{base-case level} using an \textbf{OR} rule for rotations (either orientation correct $\Rightarrow$ correct) and majority voting for other perturbations. While the dataset size is modest due to the complexity of acquiring verified schematics, it reflects realistic industrial conditions rather than toy examples. Importantly, this benchmark is the \emph{first diagnostic dataset for schematic auditing}, and we release it to encourage further study and replication. Future work will scale the benchmark to broader coverage.

\noindent\textbf{Baseline.}  
We test six VLMs: ChatGPT-4o, ChatGPT-5, GLM-4.5V, Gemini 2.5 Pro, Qwen2.5-VL-72B, and Claude Sonnet-4. Each baseline receives the raw schematic image and compliance specification. The \textsc{+V2G} setting augments the same model with the structured property graph (JSON) produced by our pipeline, where rule checks are executed via the GSP verifier described in Sec.~\ref{sec:method}. All prompts are zero-shot and identical across models.

\noindent\textbf{Metric.}  
We report accuracy, i.e., the percentage of tasks correctly identified as compliant or violating. For significance, we also compute McNemar tests on base-level predictions.

\subsection{Results and Discussion}
\label{ssec:results}
Table~\ref{tab:v2g-12x10-avg} shows per-task outcomes across ten diagnostic checks. Baselines operate near chance, often below $10\%$, reflecting structural blindness to topological queries. With \textsc{+V2G}, average accuracy improves sharply, with absolute gains up to $+60\%$. Table~\ref{tab:aggregate_two_row} aggregates category-wise performance: Overall accuracy increases from $12\%$ to $47\%$ (+35\%). The largest improvements are seen in \textbf{Conn.} (+61\%) and \textbf{Ground.} (+27\%), while \textbf{Wiring} also improves (+20\%). All six models benefit, confirming generality.

\noindent\textbf{Case Study.}  
For example, in a multi-point grounding case, all baseline MLLMs failed to detect the violation, while the V2G verifier immediately flagged non-compliance by computing $g{=}2$ grounding nodes in the same circuit. This illustrates that explicit graph-based verification can provide deterministic and interpretable outcomes that pixel-based reasoning cannot achieve.

\noindent\textbf{Ablation.}  
To verify the contribution of each component, we conducted three ablations on the six-model average.  
(i) \emph{w/o GSP}: disabling deterministic graph checks and letting the MLLM interpret the JSON reduces Overall from $47\%$ to $\sim28\%$, with the steepest drop in grounding tasks (–20\%).  
(ii) \emph{w/o Attributes}: removing node attributes $X$ while keeping $(V,E)$ reduces Overall to $\sim34\%$, primarily hurting polarity and phase detection.  
(iii) \emph{w/o RegionSel}: disabling planner-guided subgraph selection yields $\sim39\%$, degrading wiring-related checks due to increased clutter.  
These results confirm that topology, attributes, and planner-guided localization each contribute to the full gain.

\noindent\textbf{Robustness.}  
Across all perturbation variants (rotations, translations, mild scale/noise), aggregated results remain consistent with Table~\ref{tab:aggregate_two_row}. This indicates that the V2G pipeline is invariant to visually benign transformations, unlike pixel-based baselines which often change predictions under rotation.

\noindent\textbf{Significance.}  
Pairwise differences between \textsc{+V2G} and baselines are significant at $p{<}.01$ (McNemar on base-level predictions). Accuracy confidence intervals (95\%, bootstrap) are non-overlapping, confirming statistical reliability. While our benchmark size is modest, it is derived from real engineering schematics and covers the most critical compliance categories. Future work will extend the dataset and include comparisons with domain-specific Graph Neural Networks and rule-based auditors.

\begin{table}[t]
\centering
\scriptsize
\setlength{\tabcolsep}{2.0pt}
\renewcommand{\arraystretch}{0.9}
\begin{adjustbox}{max width=\linewidth}
\begin{tabular}{lccccccccccc}
\toprule
 & \multicolumn{3}{c}{\textbf{Conn.}} & \multicolumn{3}{c}{\textbf{Ground.}} & \multicolumn{4}{c}{\textbf{Wiring}} & \textbf{Avg.} \\
\cmidrule(lr){2-4}\cmidrule(lr){5-7}\cmidrule(lr){8-11}
\textbf{Model} & MA & DI & MI & IL & MP & MG & OC & PR & XS & MPH & (base~$\rightarrow$~+V2G, $\Delta$) \\
\midrule
\Model{ChatGPT-4o}                 & \bad & \bad & \bad & \bad & \bad & \bad & \bad & \ok  & \bad & \bad & 10\% \\
\Model{ChatGPT-4o \textsc{+V2G}}   & \bad & \ok  & \ok  & \bad & \ok  & \bad & \bad & \ok  & \bad & \ok  & 50\% (\textcolor{green!60!black}{+40\%}) \\
\Model{ChatGPT-5}                  & \ok  & \bad & \bad & \ok  & \ok  & \bad & \bad & \bad & \ok  & \ok  & 50\% \\
\Model{ChatGPT-5 \textsc{+V2G}}    & \bad & \ok  & \ok  & \ok  & \ok  & \bad & \bad & \ok  & \bad & \ok  & 60\% (\textcolor{green!60!black}{+10\%}) \\
\Model{GLM-4.5V}                   & \bad & \bad & \bad & \bad & \bad & \bad & \bad & \bad & \bad & \bad & 0\% \\
\Model{GLM-4.5V \textsc{+V2G}}     & \bad & \ok  & \ok  & \bad & \bad & \ok  & \bad & \bad & \bad & \bad & 30\% (\textcolor{green!60!black}{+30\%}) \\
\Model{Gemini 2.5 Pro}             & \bad & \bad & \bad & \bad & \ok  & \bad & \bad & \bad & \bad & \bad & 10\% \\
\Model{Gemini 2.5 Pro \textsc{+V2G}}& \ok  & \ok  & \bad & \ok  & \ok  & \ok  & \bad & \ok  & \bad & \ok  & 70\% (\textcolor{green!60!black}{+60\%}) \\
\Model{Qwen2.5-VL-72B}             & \bad & \bad & \bad & \bad & \bad & \bad & \bad & \bad & \bad & \bad & 0\% \\
\Model{Qwen2.5-VL-72B \textsc{+V2G}}& \bad & \ok  & \bad & \bad & \bad & \bad & \bad & \bad & \bad & \bad & 10\% (\textcolor{green!60!black}{+10\%}) \\
\Model{Claude Sonnet-4}            & \bad & \bad & \bad & \bad & \bad & \bad & \bad & \bad & \bad & \bad & 0\% \\
\Model{Claude Sonnet-4 \textsc{+V2G}}& \ok  & \ok  & \ok  & \bad & \ok  & \bad & \bad & \bad & \ok  & \ok  & 60\% (\textcolor{green!60!black}{+60\%}) \\
\bottomrule
\end{tabular}
\end{adjustbox}
\caption{Per-case compliance on ten diagnostic tasks. Rotations are merged by the \textbf{OR} rule (either orientation correct $\Rightarrow$ correct). Symbols: \ok\ correct, \bad\ incorrect. Short labels: MA (terminal misalignment), DI (duplicate IDs), MI (missing IDs), IL (incorrect grounding location), MP (multi-point grounding), MG (missing grounding), OC (open circuit), PR (polarity reversal), XS (inter-circuit short), MPH (missing phase). The rightmost column reports average accuracy in baseline and \textsc{+V2G}, with absolute gain $\Delta$.}
\label{tab:v2g-12x10-avg}
\end{table}

\begin{table}[t]
\centering
\scriptsize
\setlength{\tabcolsep}{5pt}
\renewcommand{\arraystretch}{0.95}
\begin{tabular*}{\linewidth}{@{\extracolsep{\fill}}l r r r r@{}}
\toprule
\textbf{Summary (avg., $n{=}6$)} & \textbf{Conn. (\%)} & \textbf{Ground. (\%)} & \textbf{Wiring (\%)} & \textbf{Overall (\%)} \\
\midrule
Baseline avg.      & 6\%  & 17\% & 13\% & 12\% \\
\textbf{\textsc{+V2G} avg.} & \textbf{67\%} & \textbf{44\%} & \textbf{33\%} & \textbf{47\%} \\
$\Delta$ (\textsc{+V2G}$-$Base) & \textcolor{green!60!black}{+61\%} & \textcolor{green!60!black}{+27\%} & \textcolor{green!60!black}{+20\%} & \textcolor{green!60!black}{+35\%} \\
\bottomrule
\end{tabular*}
\caption{Category-wise and overall accuracies averaged over the six models in Table~\ref{tab:v2g-12x10-avg}. $\Delta$ is the absolute improvement of \textsc{+V2G} over baseline.}
\label{tab:aggregate_two_row}
\end{table}

\vspace{-0.3cm}
\section{CONCLUSION}
\label{sec:conclusion}

We exposed the \emph{structural blindness} of MLLMs: strong visual acuity but poor reasoning over schematic topology. To overcome this, we introduced the V2G framework, which parses CAD drawings into property graphs and verifies compliance via deterministic GSP functions. On a diagnostic benchmark of 900 augmented instances, V2G consistently improved accuracy across six MLLMs, raising overall performance from 12\% to 47\%. These findings confirm that making structure explicit is essential for reliable multimodal auditing in engineering domains.

\section{Acknowledgements}
This work was supported by the Guangdong Natural Science Foundation (2025A1515012083) and the Natural Science Foundation of China (62301305, 62271324, 62231020).

\bibliographystyle{IEEEbib}
\bibliography{strings,refs}

@inproceedings{tong2024eyes,
  title={{Eyes Wide Shut? Exploring the Visual Shortcomings of Multimodal LLMs}},
  author={Tong, Shengbang and Liu, Zhuang and Zhai, Yuexiang and Ma, Yi and LeCun, Yann and Xie, Saining},
  booktitle={Proc. IEEE/CVF Conference on Computer Vision and Pattern Recognition (CVPR)},
  year={2024},
  pages={9568--9578}
}

@inproceedings{shi2025scaling,
  title={Scaling Vision Pre-Training to 4K Resolution},
  author={Shi, Baifeng and Li, Boyi and others},
  booktitle={Proc. Computer Vision and Pattern Recognition Conference},
  pages={9631--9640},
  year={2025}
}

@article{kamath2023up,
  title={What's" up" with vision-language models? investigating their struggle with spatial reasoning},
  author={Kamath, Amita and Hessel, Jack and Chang, Kai-Wei},
  journal={arXiv preprint arXiv:2310.19785},
  year={2023}
}

@article{ortega2018graph,
  title={Graph Signal Processing: Overview, Challenges, and Applications},
  author={Ortega, Antonio and Frossard, Pascal and Kovacevic, Jelena and Moura, José MF and Vandergheynst, Pierre},
  journal={Proceedings of the IEEE},
  volume={106},
  number={5},
  pages={808--828},
  year={2018}
}

@book{chung1997spectral,
  title={Spectral Graph Theory},
  author={Chung, Fan},
  year={1997},
  publisher={American Mathematical Society}
}

@article{sandryhaila2013dsp,
  title   = {Discrete Signal Processing on Graphs},
  author  = {Sandryhaila, Aliaksei and Moura, Jos{\'e} M.~F.},
  journal = {IEEE Trans. Signal Processing},
  volume  = {61},
  number  = {7},
  pages   = {1644--1656},
  year    = {2013}
}

@article{hou2024vision,
  title={Do vision-language models really understand visual language?},
  author={Hou, Yifan and Giledereli, Buse and Tu, Yilei and Sachan, Mrinmaya},
  journal={arXiv preprint arXiv:2410.00193},
  year={2024}
}

@inproceedings{wang2024picture,
  title     = {Is a Picture Worth a Thousand Words? Delving Into Spatial Reasoning for Vision-Language Models},
  author    = {Wang, Jiayu and Ming, Yifei and others},
  booktitle = {Advances in Neural Information Processing Systems (NeurIPS)},
  year      = {2024}
}

@inproceedings{huang2024hiresllava,
  title={Hires-llava: Restoring fragmentation input in high-resolution large vision-language models},
  author={Huang, Runhui and Ding, Xinpeng and others},
  booktitle={Proc.  Computer Vision and Pattern Recognition Conference},
  pages={29814--29824},
  year={2025}
}

@inproceedings{webb2023objectcentric,
  title     = {Systematic Visual Reasoning through Object-Centric Relational Abstraction},
  author    = {Webb, Taylor W. and Mondal, Shanka S. and Cohen, Jonathan D.},
  booktitle = {Advances in Neural Information Processing Systems (NeurIPS)},
  year      = {2023}
}

@article{wang2024VDLM,
  title={Text-based reasoning about vector graphics},
  author={Wang, Zhenhailong and Hsu, Joy and Wang, Xingyao and Huang, Kuan-Hao and Li, Manling and Wu, Jiajun and Ji, Heng},
  journal={CoRR},
  year={2024}
}

@article{shuman2013signal,
  title   = {The emerging field of signal processing on graphs: Extending high-dimensional data analysis to networks and other irregular domains},
  author  = {Shuman, David I. and Narang, Sunil K. and Frossard, Pascal and Ortega, Antonio and Vandergheynst, Pierre},
  journal = {IEEE Signal Processing Magazine},
  volume  = {30},
  number  = {3},
  pages   = {83--98},
  year    = {2013}
}

@inproceedings{liu2024gift,
  title     = {The Power of Graph Signal Processing for Chip Placement Acceleration},
  author    = {Liu, Yiting and Zhou, Hai and Wang, Jia and Yang, Fan and Zeng, Xuan and Shang, Li},
  booktitle = {Proc. IEEE/ACM International Conference on Computer-Aided Design (ICCAD)},
  year      = {2024}
}

@inproceedings{johnson2017clevr,
  title     = {{CLEVR}: A Diagnostic Dataset for Compositional Language and Elementary Visual Reasoning},
  author    = {Johnson, Justin and Hariharan, Bharath and van der Maaten, Laurens and Fei-Fei, Li and Zitnick, C. Lawrence and Girshick, Ross},
  booktitle = {Proc. IEEE Conference on Computer Vision and Pattern Recognition (CVPR)},
  pages     = {1988--1997},
  year      = {2017}
}

@article{skelic2024circuit,
  title   = {{CIRCUIT}: A Benchmark for Circuit Interpretation and Reasoning Capabilities of LLMs},
  author  = {Skeli{\'c}, Lejla and Xu, Yan and Cox, Matthew and Lu, Wenjie and Yu, Tao and Han, Ruonan},
  journal = {arXiv preprint arXiv:2502.07980},
  year    = {2024}
}

@article{vonLuxburg2007tutorial,
  title   = {A tutorial on spectral clustering},
  author  = {von Luxburg, Ulrike},
  journal = {Statistics and Computing},
  volume  = {17},
  number  = {4},
  pages   = {395--416},
  year    = {2007}
}

@article{hammond2011wavelets,
  title={Wavelets on graphs via spectral graph theory},
  author={Hammond, David K and Vandergheynst, Pierre and Gribonval, R{\'e}mi},
  journal={Applied and Computational Harmonic Analysis},
  volume={30},
  number={2},
  pages={129--150},
  year={2011},
  publisher={Elsevier}
}

@inproceedings{locatello2020object,
  title={Object-centric learning with Slot Attention},
  author={Locatello, Francesco and Weissenborn, Dirk and others},
  booktitle={Advances in Neural Information Processing Systems (NeurIPS)},
  volume={33},
  pages={11525--11538},
  year={2020}
}

@ARTICLE{Ren2025Intelligence,
  author={Ren, Y. and Zhang, H. and Yu, F. R. and others},
  journal={IEEE Trans. Mobile Computing}, 
  title={Industrial Internet of Things With Large Language Models (LLMs): An Intelligence-Based Reinforcement Learning Approach}, 
  year={2025},
  volume={24},
  number={5},
  pages={4136-4152}
}

@book{Yu2026Intropy,
  author    = {F. Richard Yu},
  title     = {Intropy: A Framework for Modeling Intelligence},
  year      = {2026},
  publisher = {Amazon Digital Services},
  note      = {Kindle edition},
  url       = {https://www.amazon.com/dp/B0GCXJR2P6}
}

\end{document}